\documentclass{article}

\usepackage{PRIMEarxiv}

\usepackage[utf8]{inputenc} 
\usepackage[T1]{fontenc}    
\usepackage{hyperref}       
\usepackage{url}            
\usepackage{booktabs}       
\usepackage{amsfonts}       
\usepackage{nicefrac}       
\usepackage{microtype}      
\usepackage{lipsum}
\usepackage{fancyhdr}       
\usepackage{graphicx}       
\graphicspath{{media/}}     
\usepackage{listings}
\title{A Logical Approach to Criminal Case Investigation}

\author{
  Takanori Ugai \\
  Fujitsu Limited \\
  Kanagawa, Japan\\
  \texttt{ugai@fujitsu.com} \\
   \And
  Yusuke Koyanagi\\
  Fujitsu Limited \\
  Kanagawa, Japan\\
   \And
  Fumihito Nishino \\
  Fujitsu Limited \\
  Kanagawa, Japan\\
}

\begin{document}
\maketitle


\begin{abstract}
XAI (eXplanable AI) techniques that have the property of explaining the reasons for their conclusions, i.e. explainability or interpretability, are attracting attention.
XAI is expected to be used in the development of forensic science and the justice system. In today's forensic and criminal investigation environment, experts face many challenges due to large amounts of data, small pieces of evidence in a chaotic and complex environment, traditional laboratory structures and sometimes inadequate knowledge.
All these can lead to failed investigations and miscarriages of justice. In this paper, we describe the application of one logical approach to crime scene investigation.
The subject of the application is ``The Adventure of the Speckled Band'' from the Sherlock Holmes short stories.
The applied data is the knowledge graph created
for the Knowledge Graph Reasoning Challenge.
We tried to find the murderer by inferring each person with the motive, opportunity, and method.
We created an ontology of motives and methods of murder from dictionaries and dictionaries, added it to the knowledge graph of ``The Adventure of the Speckled Band'', and applied scripts to determine motives, opportunities, and methods.
\end{abstract}



\maketitle

\section{Introduction}
XAI (eXplanable AI) techniques that have the property of explaining the reasons for their conclusions, i.e. explainability or interpretability, are attracting attention.
XAI is expected to be used in the development of forensic science and the justice system. In today's forensic and criminal investigation environment, experts face many challenges due to large amounts of data, small pieces of evidence in a chaotic and complex environment, traditional laboratory structures and sometimes inadequate knowledge.
All these can lead to failed investigations and miscarriages of justice. Artificial intelligence (AI), such as machine learning and deep learning, is a weapon in the arsenal to combat these difficulties. In many areas of forensics, neural networks and logical reasoning are expected to produce error-free, objective and reproducible conclusions\cite{Sanskriti_Kadiyan_2021}.
In this paper, we describe the application of one logical approach to crime scene investigation.
The subject of the application is ``The Adventure of the Speckled Band''\cite{doyle2016adventure} from the Sherlock Holmes short stories.
The applied data is the knowledge graph created
for the Knowledge Graph Reasoning Challenge
\footnote{\url{http://challenge.knowledge-graph.jp/}}\cite{10.1007/978-3-030-41407-8_2}.
We tried to find the murderer by inferring the person who has the motive, opportunity and method.
As shown in Fig. \ref{Fig5}, we created an ontology of motives and methods of murder from dictionaries, added it to the knowledge graph of ``The Adventure of the Speckled Band'', and applied scripts to determine motives, opportunities, and methods.
From these scripts, we obtained characters with motives, characters with opportunities to kill, and characters who could execute the methods of killing deduced from the situation at the crime scene. We made the final overall judgment manually based on the assessment of motive, opportunity, and method.

The contribution of this study can be summarized as follows.
\begin{itemize}
\item We developed and published a system to solve murder investigation. The system explained who did it, why and how.
\item We developed and published motivation ontology and means ontology to help the investigation.
\item We developed and published rules with semantic technology (SPARQL and SHACL) to represend the process of murder investigation.
\item We showed a use case to apply our system to a Sherlock Homes short Story.
\end{itemize}

Section 2 describes related works.
In Section 3, we describe the knowledge graph to which we applied our investigation process. The knowledge graph is defined and 
created by the organisers of the knowledge graph reasoning challenge.
Section 4 describes our investigation process.
Finally, in Section 5, we give a summary and discuss future works.

\section{Related Works}
There are some studies of ontology model to represent crime information.
Jalil\cite{Jalil2017} developed an ontology model using the selected semantic modelling tool to represent the crime information with the well-defined classes and relationships. This ontology model helps to save the investigation officer’s effort in aiming and targeting the possible suspect within the shortest time interval.
Elezaj\cite{Elezaj2019} presents a knowledge graph-based framework, an outline of a framework designed to support crime investigators solve and prevent crime, from data collection to inferring digital evidence admissible in court.
Vakaj\cite{Vakaj2017} developed the SMONT ontology which gives support to the process of crime investigation and prevention. The SMONT ontology covers specific data about the crime, digital evidence obtained from the online social network,
Srimukh\cite{Srimukh2020} deals with the working principle and the construction  of an ontology-based extensively on organised crime and describes the structure of the ontology they created and also by validating the ontology via an online ontology evaluating tool. 
Chabot\cite{CHABOT2014S95} propose a methodology, supported by theoretical concepts, that can assist investigators through the whole process including the construction and the interpretation of the events describing the case. The proposed approach is based on a model that integrates experts' knowledge from digital forensics and software development fields to allow a semantically rich representation of events related to the incident.
Elsayed\cite{Elsayed2018} proposed SCIS (Semantic Crime Investigation System), which
supports investigator to take a certainty decision based on the result of the fusion. The solution proposed in this paper
(SCIS) used Ontology re-engineering to create Crime Universal Ontology which includes kinesics Ontology. Also
SCIS applied sentiment analysis and image processing techniques.

The contribution of this paper is to formulate the process of case investigation and to give an example of its application to a case represented in a knowledge graph.

\section{Crime Scene Knowledge Graph}
\subsection{Knowledge Graph Schema}
The Knowledge Graph Reasoning Challenge aims to bring together methods for inference and estimation from a wide range of 
professionals, and to objectively evaluate, classify and systematise them.
In 2018, the first year of the challenge, the organisers published a knowledge graph based on Sherlock Holmes' short mystery story
``The Adventure of the Speckled Band'', the criminal case, background and characters are turned into knowledge.

\begin{figure}[ht]
\centering
\includegraphics[width=1.0\linewidth]{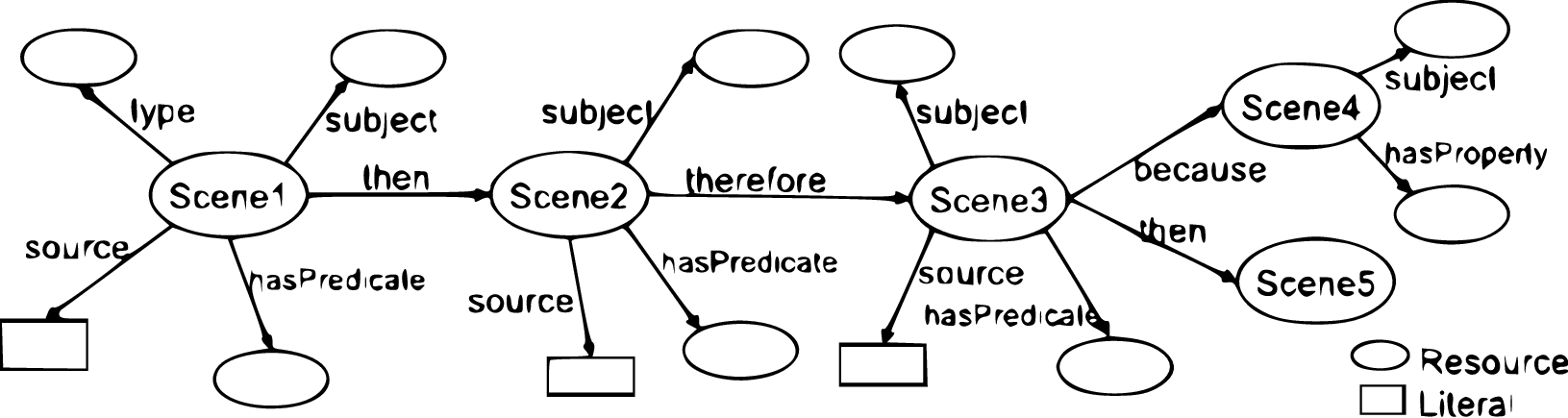}
  \caption{Architecture of the knowledge graph}\label{Fig0}
\end{figure}

The knowledge graph provided by the organisers is designed to represent temporal, causal and probabilistic relationships that reflect the real world.
Resource Description Framework (RDF) represents the different situations in a unified and computer-processable form and can be searched by SPARQL, a standard query language for graph DB.
The schema of the knowledge graph is based on the division of a set of contents into scenes, and the graphical representation of the contents of each scene and the relationships between scenes.
Fig. \ref{Fig0}\cite{10.1007/978-3-030-41407-8_2} shows the overall image of the knowledge graph.

\begin{figure}[ht]
\centering
\includegraphics[width=1.0\linewidth]{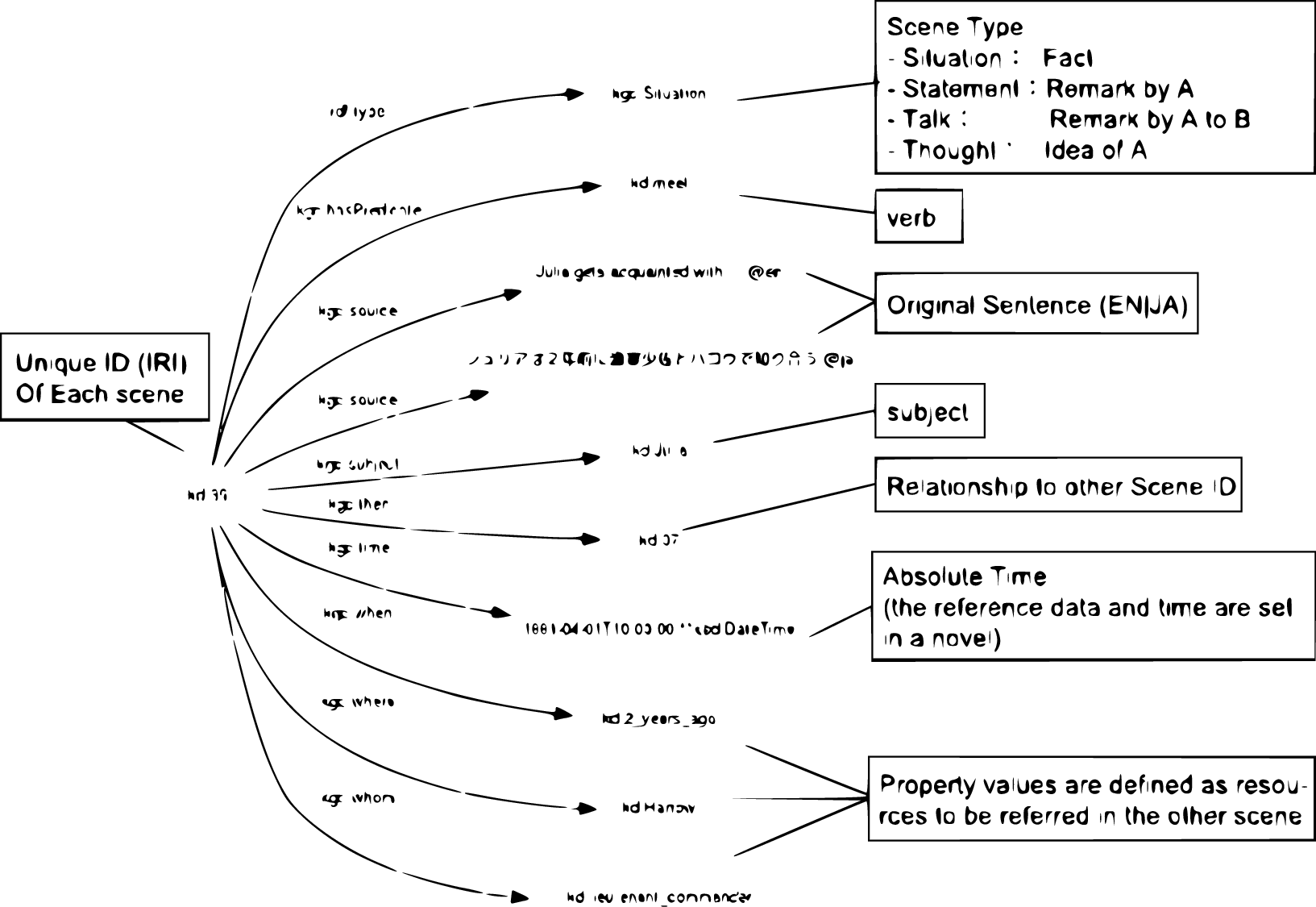}
  \caption{Example of a scene graph}\label{Fig2}
\end{figure}

The following basic properties are used to describe each scene.
Unlike the general <subject, predicate, object> format, this property uses the scene ID as the subject, to summarise the information associated with the scene.
Fig. \ref{Fig2}\cite{10.1007/978-3-030-41407-8_2} shows an example of scene description.

\begin{itemize}
\item subject:The person or thing that is the subject of the description of the scene
\item hasPredicate: A predicate describing the content of the scene
\item hasProperty: The nature of the person or thing that is the subject
\item Objectives with scene details: whom, where, when, what, how etc.
\item Relationships between scenes: then,if, because, etc.
\item time: The absolute time the scene took place (xsd:DateTime)
\item source: The original text of the scene (Literals in English and Japanese)
\end{itemize}

In addition, to distinguish between facts/situations and statements/thoughts, the person corresponding to A is described as the source of the information using the kgcc:infoSource property if it is A's (to B) statement(Scene ID type is Statement or Talk)  or A's thought (Scene ID type is Thought).
In addition, to express the AND and OR relations between subjects and targets when there are multiple subjects and targets, the AND case is defined by describing multiple triples with the same property, and the OR case is expressed by describing a resource that represents a ``combination of ORs'' as an instance of the ORobj type.
In the case of OR, a resource representing an ``OR combination'' is described as an instance of type ORobj. From this resource, multiple resources that are the target of OR are described via the kgcc:orTarget property.
In addition, we introduced NotAction and CanNotAction type classes as subclasses of Action to handle negation of predicates (not, not possible).
Negative predicates are described as instances of these classes.
At the same time, they are connected to affirmative predicates (Action type) by the kgcc:Not and kgcc:canNot properties.

\subsection{The Adventure of the Speckled Band}
The Adventure of the Speckled Band is the eighth Sherlock Holmes short story originally published in Strand Magazine in February 1892.
The synopsis is as follows.
Helen Stoner tells Holmes and Watson about her life in the secluded house, the nature of her stepfather and the day her twin sister Julia died.
Two years ago Julia, like Helen now, was about to be married. Two years ago, Julia was about to be married, as is Helen now, but she died after screaming in her sleep and leaving behind the cryptic words ``a speckled band''.
The autopsy and subsequent investigations did not reveal anything suspicious, and the cause of death remained unclear for many years. However, Helen never forgot her sister's strange behaviour and the words she left behind.
When Helen's bedroom is being renovated, she is given her sister's bedroom at short notice. As she sleeps, she hears the ``whistling sound'' that her sister used to make before she died.
Fearing that she might end up like Julia, she comes to Baker Street for help.
After hearing of her financial situation and the eccentricities of her father-in-law, Dr. Grimsby Roylott, Holmes decided that time was of the essence. He immediately goes to the scene to investigate.

\section{Investigation Process}
To deduce the culprit, various kinds of evidence (physical evidence and circumstances), such as the circumstances of the damage, the appearance of the victim and the suspect, and their relationships, are accumulated and compared with various knowledge related to the crime. Multiple analyses are carried out to narrow down the suspects and deduce the culprit.

The clues for the identification of the perpetrator include
\begin{enumerate}
\item the analysis of motive
\item the analysis of opportunity
\item the analysis of means
\item the analysis of behaviour and knowledge
\end{enumerate}

\begin{figure*}[ht]
\centering
\includegraphics[width=1.0\linewidth]{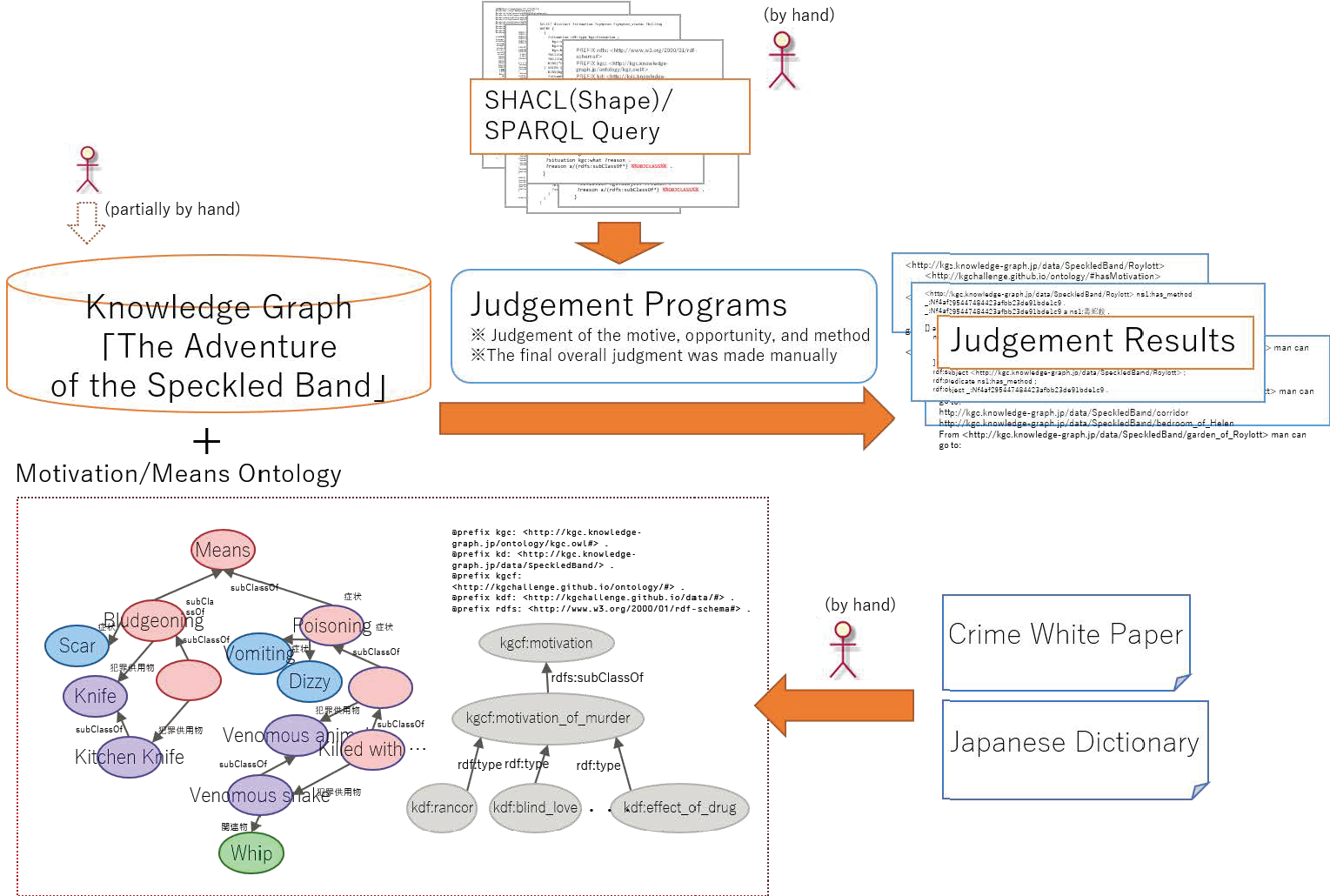}
  \caption{Overall structure of the reasoning system}\label{Fig5}
\end{figure*}

(4) analyses of who said something that only the perpetrator could know, who knew when and where the victim was, and who arranged the crime to take advantage of the perpetrator.
In this paper, we tried to find the murderer by inferring the person who has the motive, opportunity and method.
As shown in Fig. \ref{Fig5}, we created an ontology of motives and methods of murder from dictionaries and dictionaries, added it to the knowledge graph of ``The Adventure of the Speckled Band'', and applied scripts to determine motives, opportunities, and methods.
From these scripts, we obtained characters with motives, characters with opportunities to kill, and characters who could execute the methods of killing deduced from the situation at the crime scene. Based on motive, opportunity, and method, we made the final overall judgment manually.

\subsection{Created Ontologies}
In this section we describe the knowledge needed to deduce the perpetrator, the general knowledge we have developed and the schema we have used to describe the crime and the situation of the relationships.

One of the clues for estimating the criminal is crime motives.
The White Paper on Crime contains statistics on crime motives, and we mainly listed the possible motives of crime from the White Paper.
Related information to the motive of the crime is the inter-agent relationship.
The ontologies (vocabularies) describing inter-agent relationships are FOAF (Friend of Friend) {footnote{\url{http://xmlns.com/foaf/spec/}}}, SIOC (Semantically-Interlinked Online Communities) {\footnote{\url{https://www.w3.org/Submission/sioc-spec/}}, BIO (A vocabulary for biographical information) {\footnote{\url{ http://vocab.org/bio/}}, genont/srcont\footnote{\url{https://www.zandhuis.nl/sw/genealogy/}}, used in genealogy, RELATIONSHIP(A vocabulary for describing relationships between people)\footnote{\url{http://vocab.org/relationship/}} We decided to use a usable vocabulary that fits our purpose.
In this study, we used AgRelOn (An Agent Relationship Ontology)\footnote{\url{https://d-nb.info/standards/elementset/agrelon}}to describe human relationships in terms of inheritance. In this paper, we describe the relationships based on AgRelOn (An Agent Relationship Ontology).
As for the means of killing, since many words in Japanese indicate the means of killing, such as ``poisoning'', ``bludgeoning'', and ``strangulation'' simultaneously, we first listed various ways of killing based on the backward matching of ``kill'' in the Japanese language dictionary.
In the Japanese dictionary, the word's meaning is described as ``to kill by beating with a blunt instrument'', but we described these methods in detail as structured data with attributes (offering, use, place, object) and values.
An offering is an object that is directly used or intended to be used for the crime.
For example, the meaning of the Japanese word for ``strangulation'' is ``to kill by strangling the neck with the hands'', which is described as ``strangulation offering: none, use: part of the body, place: neck, action: strangulation, killing''.
We also described the symptoms before death and the traces left on the corpse.
Since this case was a poisoning, we also described the means of poisoning, the type of poison, and the poisoned animals.

\subsection{Motivation}
We developed an ontology of motives for murder by referring to the White Paper on Crime, describing the situations in which motives for murder occur, and creating a rule for inferring the relevant characters.
By adding the ontology of motives for murder to the given knowledge graph and applying the inference rules with motives, we identified the characters in the situations with motives for murder. We identified the characters that the characters wanted to kill.
An ontology of motives for murder was created using the White Paper on Crime as a reference, as shown in Fig. \ref{Fig8}.
The motives for murder were classified into nine categories: rancour, blind love, the effect of drug, self-expression, defence/self-defence, obedience/compliance, creed/belief, appropriation for living, and appropriation for entertainment.

\begin{figure}[ht]
\centering
\includegraphics[width=0.4\textwidth]{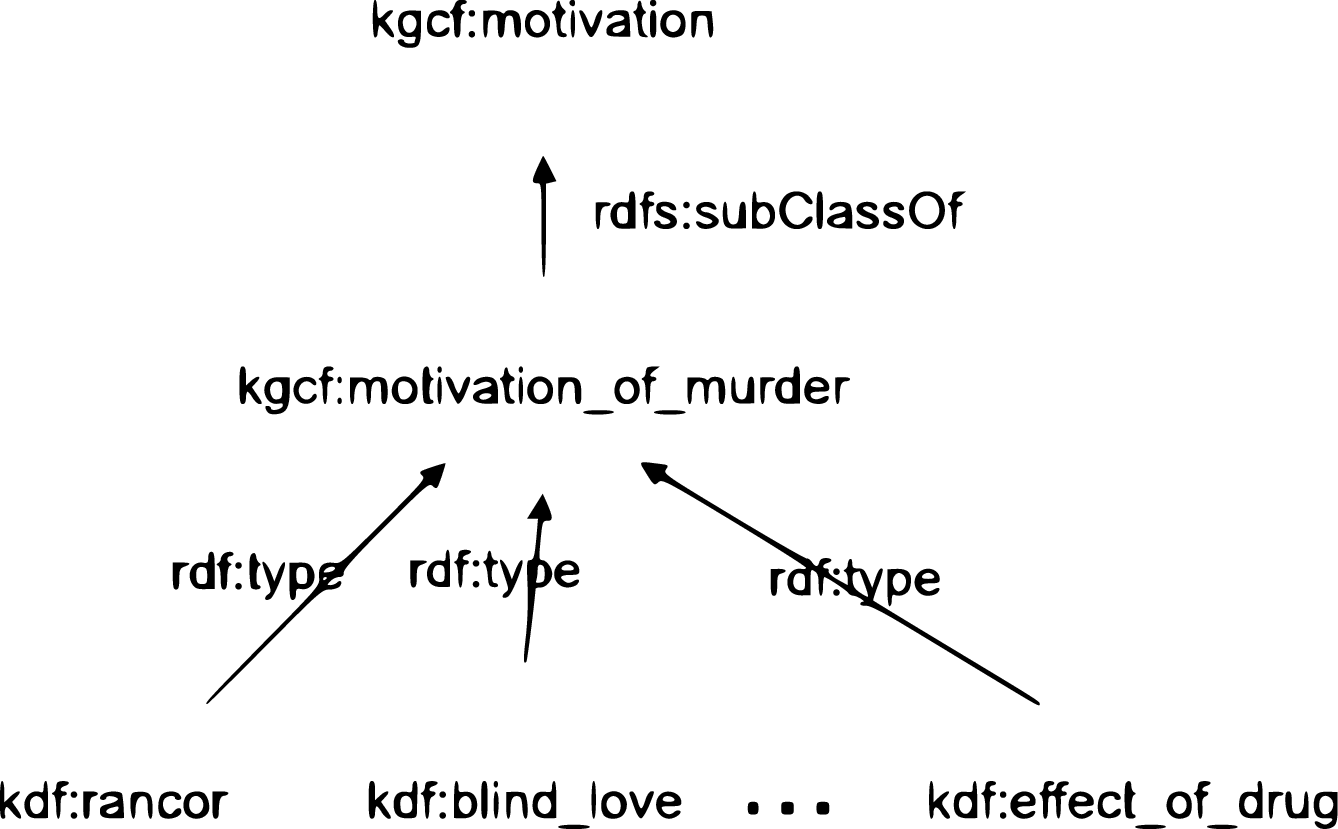}
  \caption{An ontology of motives for murder}\label{Fig8}
\end{figure}

SHACL(Shapes Constraint Language)\footnote{\url{https://www.w3.org/TR/shacl/}} is a declarative language for RDF constraints, and a formal mechanism to detect and describe violations of such constraints.
It became a W3C recommendation in 2017. SHACL groups constraints in so-called
``shapes'' to be verified by specific nodes of the graph under validation, and such that shapes may reference each other.
Fig. \ref{Fig9} is a description that assigns information about the motivation for grudges to characters in SHACL who have lost someone close to them and who are predicted to be the cause of the death.
Using the SHACL system, we obtained the following three results.

\begin{figure}[ht]
\centering
\includegraphics[width=0.5\textwidth]{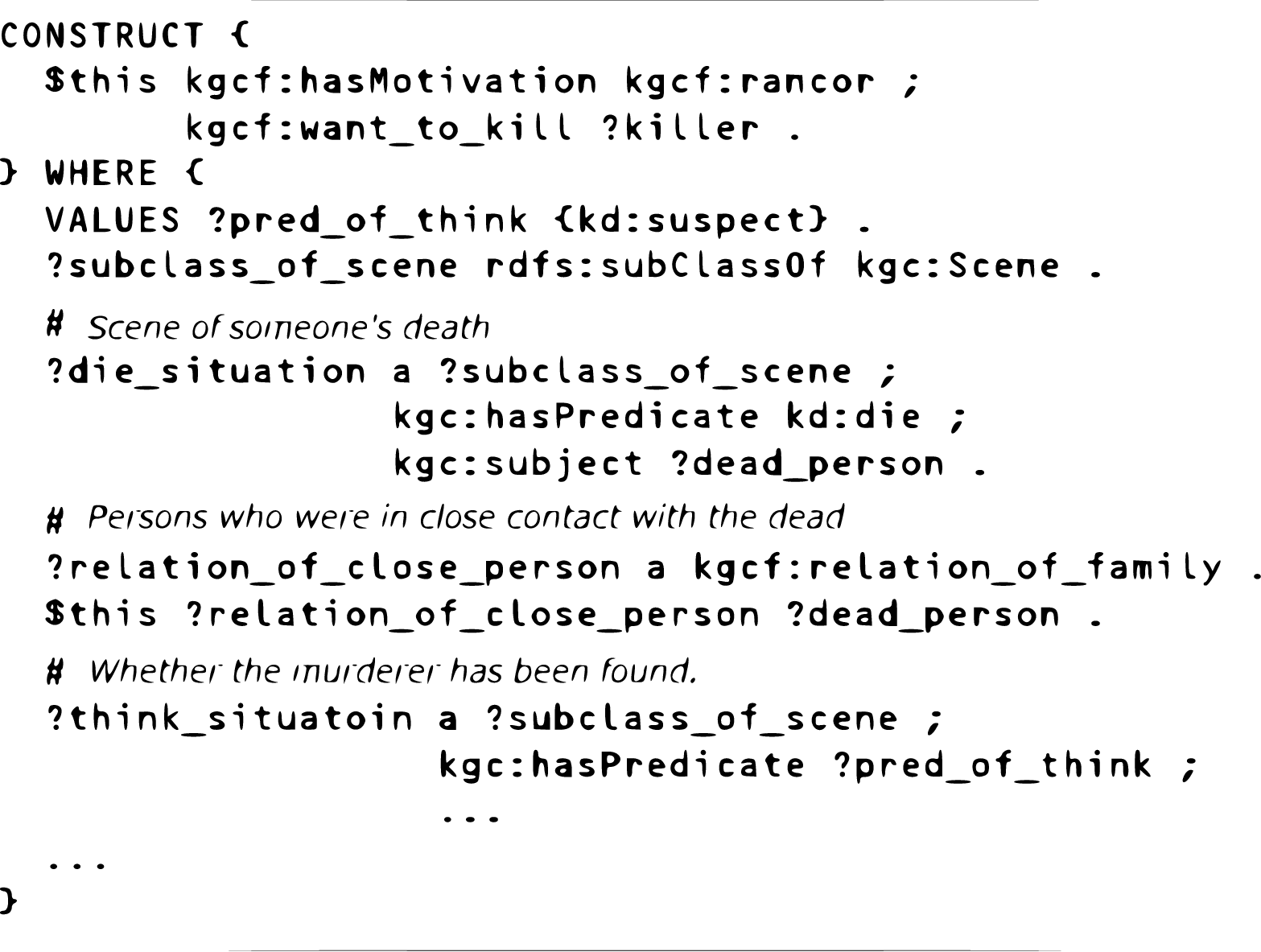}
  \caption{Rules for inferring who has a grudge motive}\label{Fig9}
\end{figure}

\begin{enumerate}
\item Roylott could kill Julia and Helen for money.
\item The villagers may kill Roylott in self-defence.
\item Helen may kill Roylott in self-defence.
\end{enumerate}

\subsection{Opportunities}
The basic principle is that there is an opportunity for someone to have been in Julia's room on the night of the incident. We divide the inference into two parts: a part for inferring temporal connections, inferring where each person was at the time of the incident, and a part for inferring spatial connections, identifying the spatial connections in which they could move.

The location of the characters at the time of the incident is deduced by retrieving the information of the scenes that have the same time as the incident, as shown in Fig.\ref{Fig10}, and excluding the scenes that arrive at the same time but are deduced to be after the incident by the property ``then''.
This inference relies on the sequence of events after the incident shown simultaneously in the provided knowledge graph.

\begin{figure}[ht]
\centering
\includegraphics[width=0.6\textwidth]{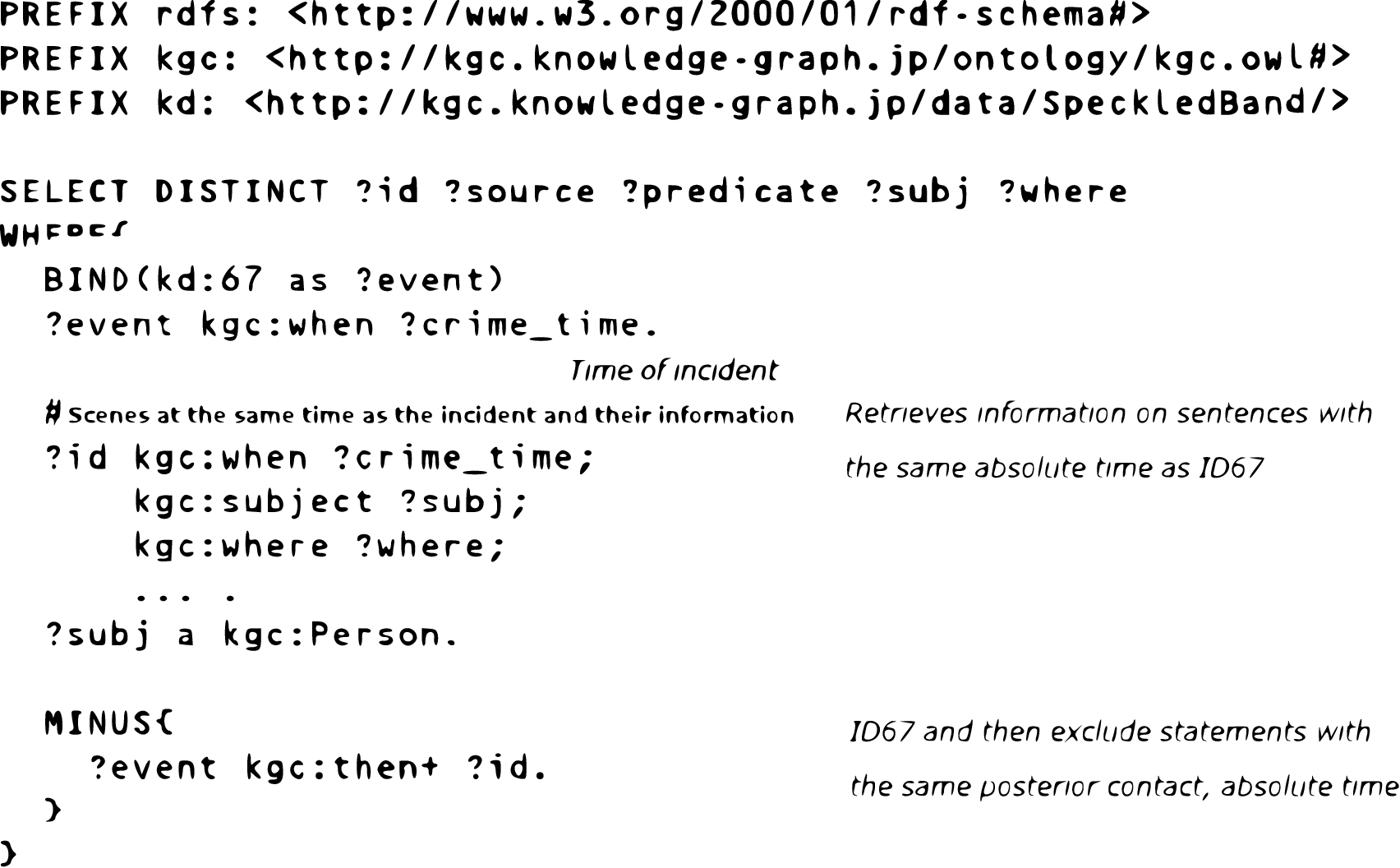}
  \caption{Rules for inferring where the characters were at the time of the incident}\label{Fig10}
\end{figure}

By this reasoning, we can infer that the five characters who were near the crime scene when it happened were in the following places
\begin{enumerate}
\item Julia was in her bedroom.
\item Helen was in her bedroom.
\item Roylott was in her bedroom.
\item Roma was in the garden.
\end{enumerate}

The next step is to deduce whether Helen, Roylott and Roma, other than the murdered Julia, can get to the bedroom where Julia is.
We list the connections connected by holes as in Fig.\ref{Fig11}, describe the connections that are not passable as in Fig.\ref{Fig12}, and subtract the remaining connections as passable.
As a result, we deduced that a person could not have passed to Julia's bedroom from each of these places.

\begin{figure}[ht]
\centering
\includegraphics[width=0.6\textwidth]{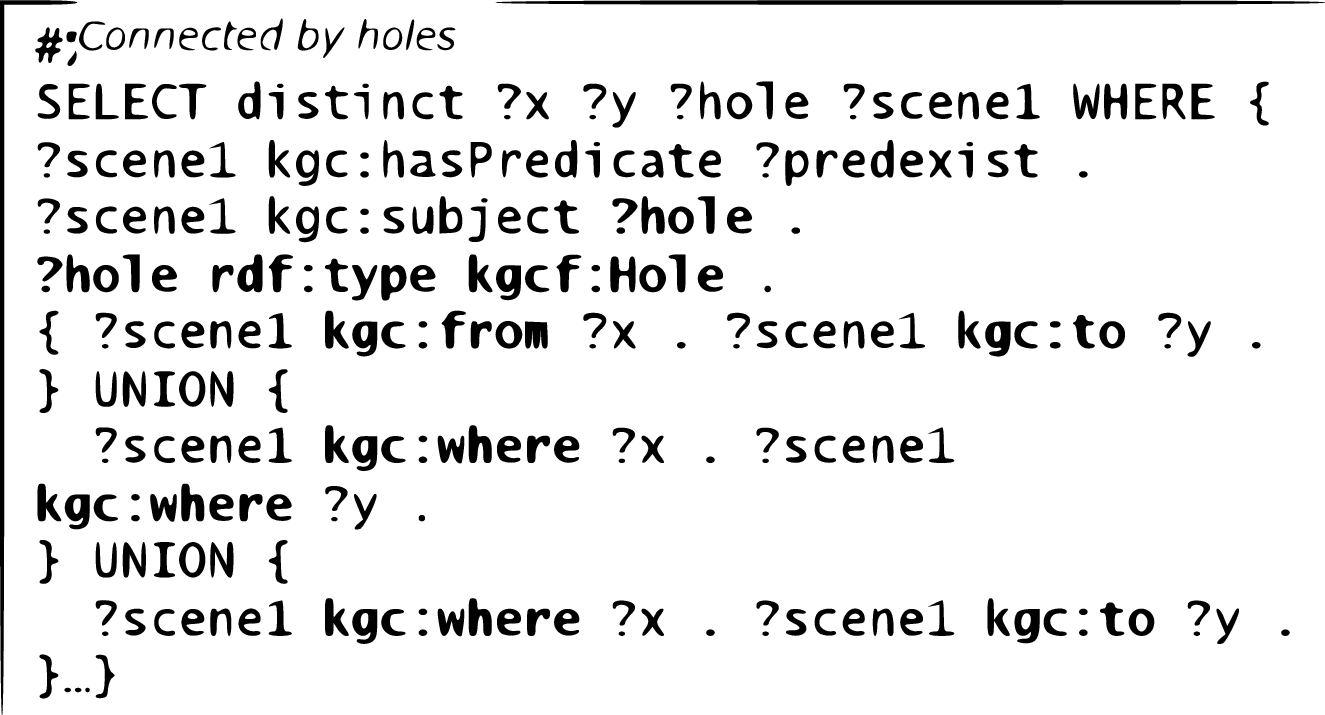}
  \caption{Rules for inferring places connected by holes}\label{Fig11}
\end{figure}

\begin{figure}[ht]
\centering
\includegraphics[width=0.5\textwidth]{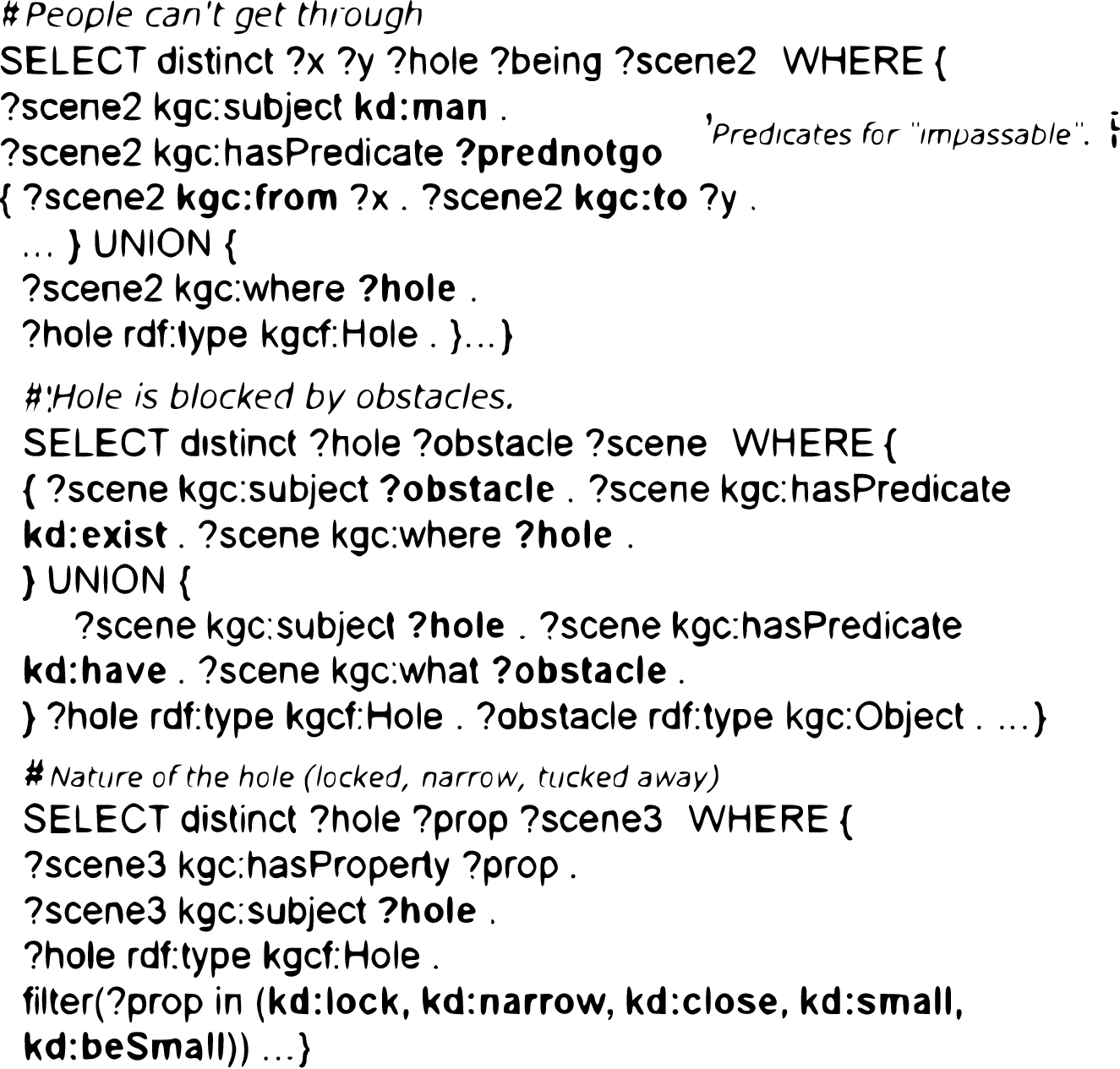}
  \caption{Rules for inferring connections that people can't get through}\label{Fig12}
\end{figure}

\subsection{Method}
The basic policy is to detect ``the person who satisfies the condition to carry out the killing method'' against ``the killing method that matches the situation of the victim and the scene''.
We developed a ``murder method ontology''. We realised two parts: one is to narrow down the murder method based on the victim and the scene on the night of the crime. The other is to derive the person who satisfies the necessary conditions to execute the narrowed murder method.
We defined the possession of a necessary object (a crime offering) as a condition for executing a killing method.
In addition, the person who has something related to the object (related object) is the person who is likely to satisfy the condition to carry out the method.

\begin{figure}[ht]
\centering
\includegraphics[width=0.7\textwidth]{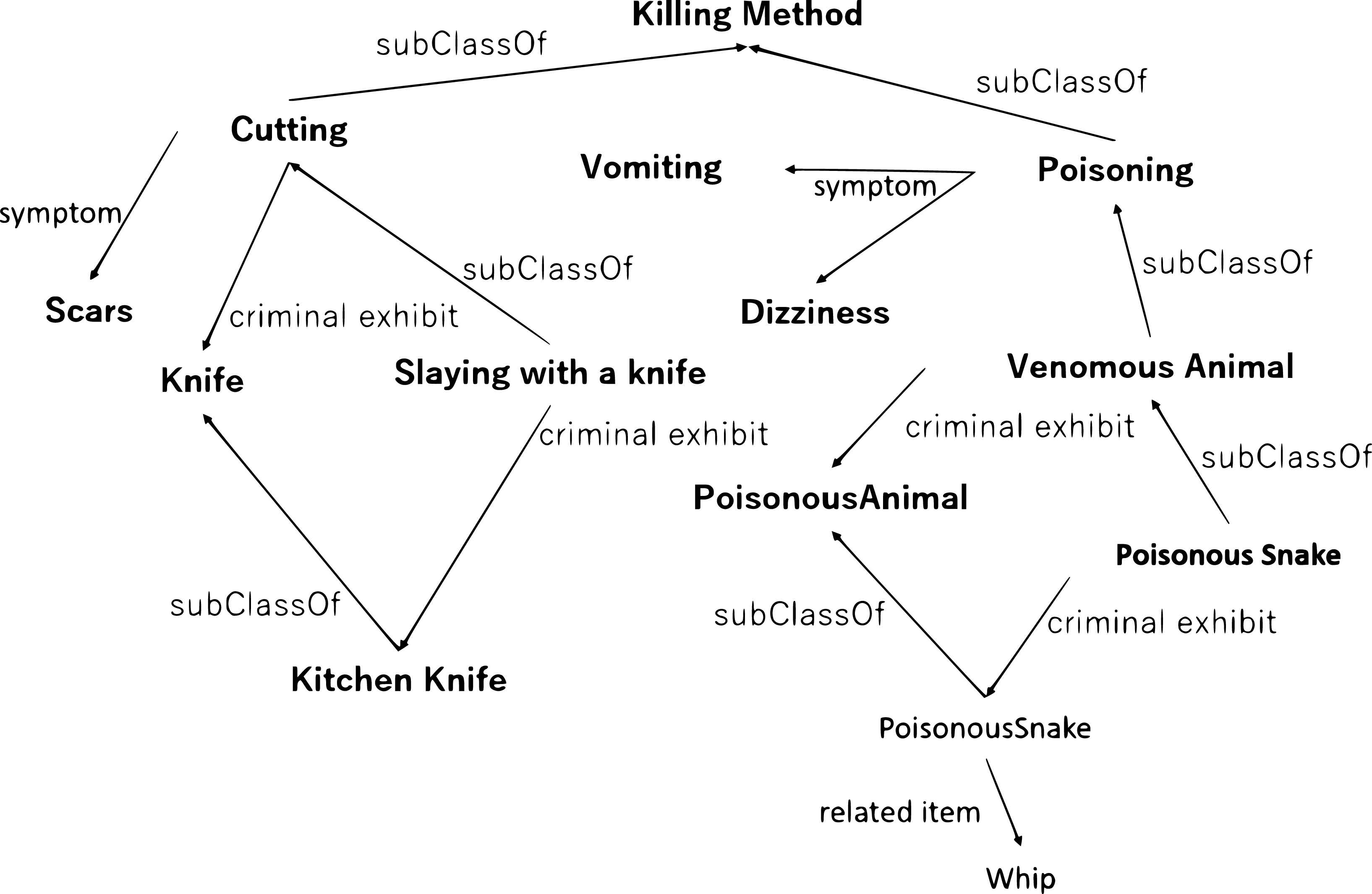}
  \caption{Part of the killing method ontology}\label{Fig13}
\end{figure}

Fig.\ref{Fig13} shows a part of the ``Killing Method Ontology''.
For each killing method class, the hierarchical relationship with other classes, the effect on the victim, the criminal object and the related object are defined.
There are 17 subclasses of killing methods, which are listed from the Web and dictionary research.
We have defined 17 subclasses of killing methods listed from the Web and dictionary research, and six subclasses.
The mapping between each class of the ontology and each instance of the knowledge graph of the inference challenge was made manually.
Fig.\ref{Fig14} shows a SPARQL query to narrow down the murder methods by comparing the victim's appearance on the day of the incident with the ontology.
This query leads to the inference that the method of murder was poisoning and that the reasons for the symptoms were ``dizziness'', ``pale'' and ``no scars''.
Fig.\ref{Fig15} is a query to retrieve the person who owns an instance of a given class.
From the given list of killing method classes, we extract the crime offerings and related objects and replace ``\%\%OBJCLASS\%\%'' in the query.
This process yielded the following results from the entire knowledge graph.

\begin{figure}[ht]
\centering
\includegraphics[width=0.6\textwidth]{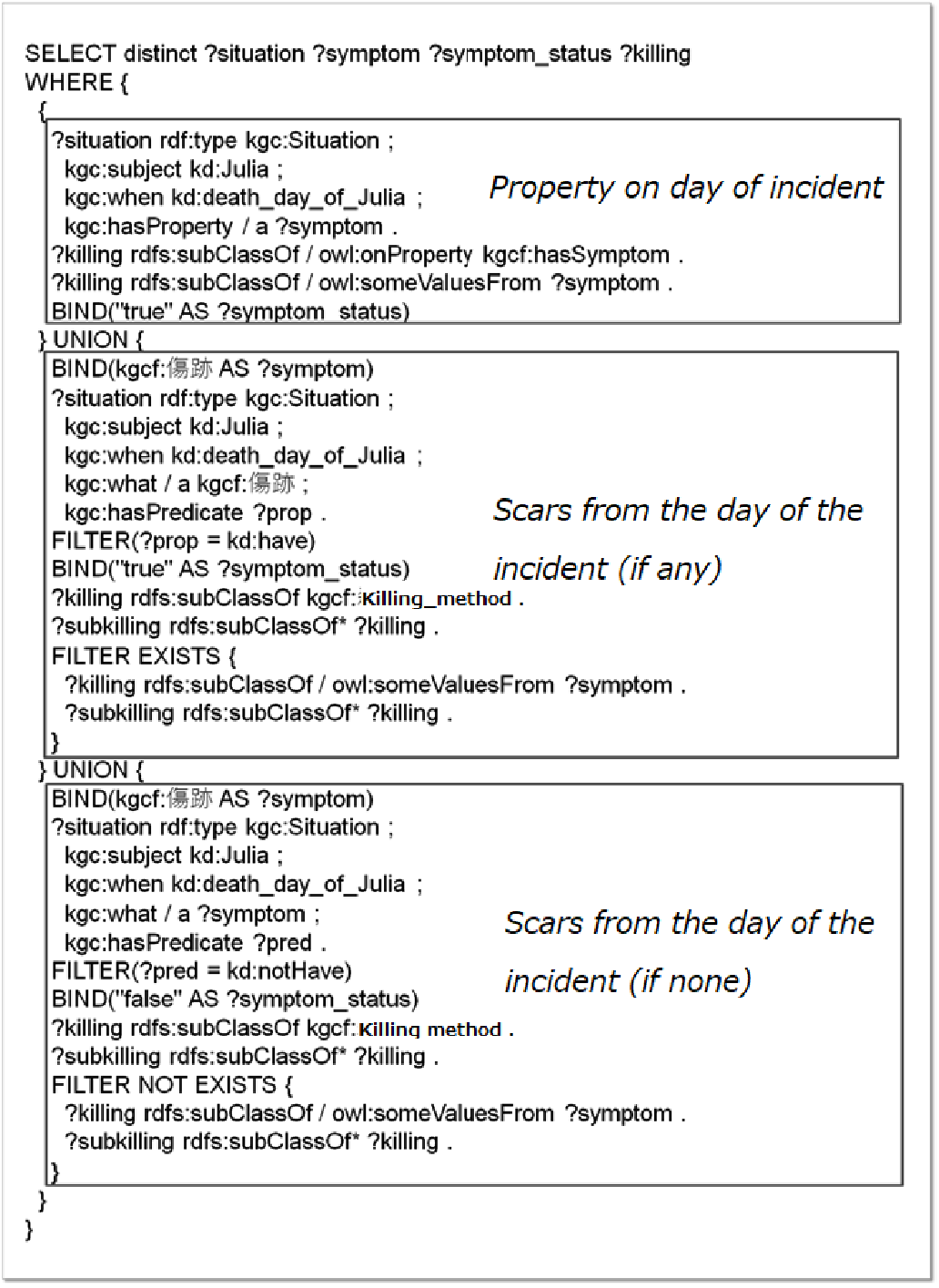}
  \caption{Query to refine the killing method}\label{Fig14}
\end{figure}

\begin{figure}[ht]
\centering
\includegraphics[width=0.6\textwidth]{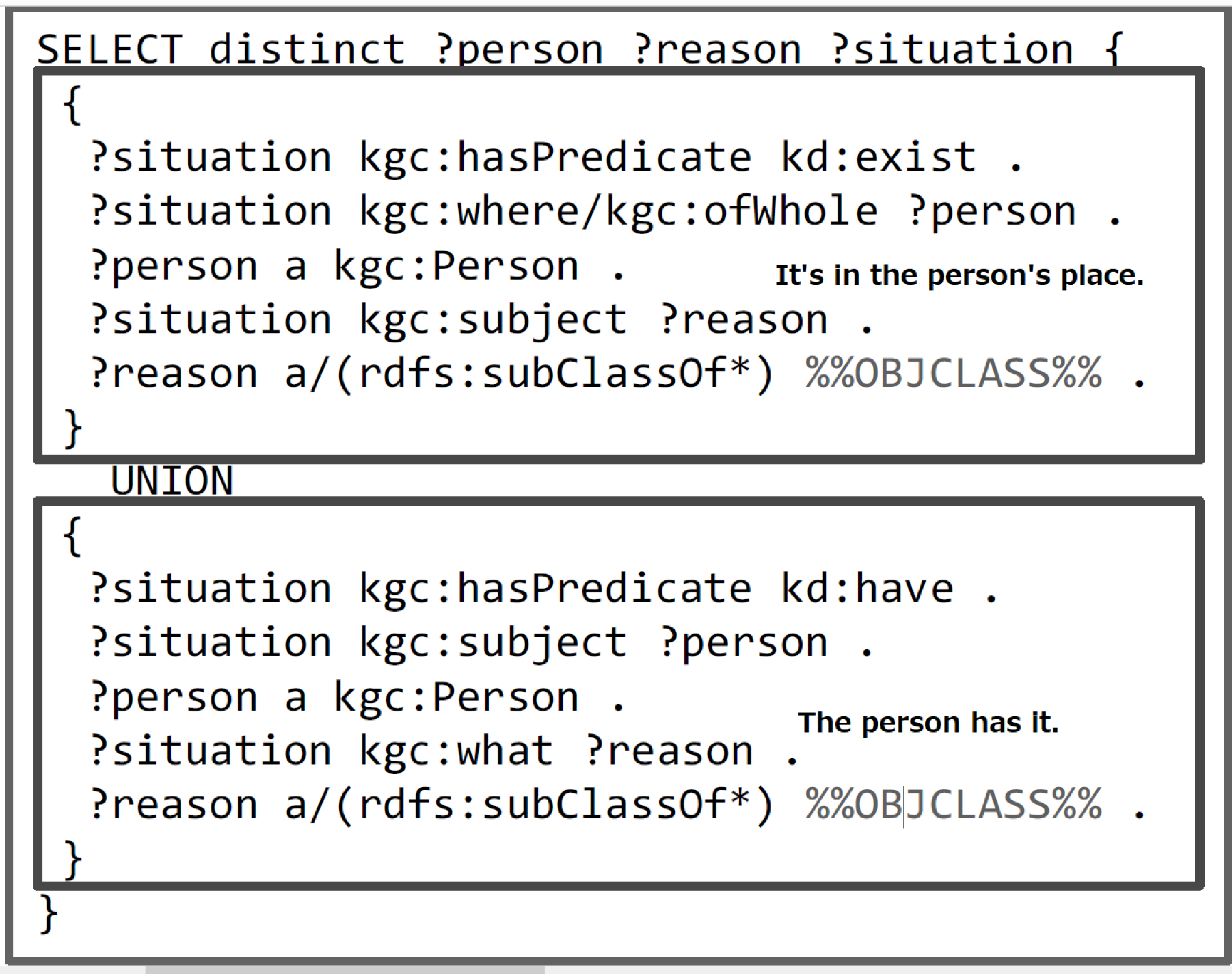}
  \caption{Query to extract persons in possession of criminal offerings and related items}\label{Fig15}
\end{figure}

\begin{enumerate}
\item Viper-killing is theorised as a viable means of killing Roylott.
The reason for this is the whip, which was in Roylott's room.
\item Poisoned animals are inferred as a viable means of killing Roylott.
The reason for this is the whip, which was in Roylott's room.
\end{enumerate}

In this study, we considered only the possession of an object as a condition for realising the method of killing, but various shapes other than possession can exist.
In addition, the relationship between criminal objects and related objects is represented by triples, which indicate the presence or absence of a simple relationship.
In fact, these relations can be of various kinds.
There is room to improve the method to deal with the variety of "conditions of the killing method" and "relations between objects", and to automate the mapping between "classes defined in the ontology" and "objects in the knowledge graph".

\subsection{Comprehensive assessment}
The judgments so far can be summarised as follows.
\begin{itemize}
\item Three motivated persons were identified. It was deduced that Roylott could kill Julia and Helen for ``greed'', the villagers could kill Roylott for ``self-defence'' and Helen could kill Roylott for ``self-defence''.
\item In terms of persons of opportunity, it was deduced that Julia was in Julia's bedroom, Helen in Helen's bedroom, Roylott in Roylott's bedroom and Roma in the garden.
However, it was not possible for a person to pass to Julia's bedroom from each of these places.
\item From the point of view of a person with a method, it was inferred from Julia's appearance at the time of her death and the fact that Roylott had a whip that Roylott ``killed her with a poisonous snake'' or ``killed her with a poisonous animal''.
\end{itemize}

From the above, we infer that Roylott killed Julia with a poisonous snake for the sake of money.

It should be noted that if we use the 90\% knowledge graph from the beginning of the case as a whole, excluding the part where Holmes reveals his story, we do not know that there was a whip in Roylott's room, so we cannot infer that Roylott had the means to kill her.

The results for motive and opportunity are unchanged. Furthermore, when Holmes used the 75\% knowledge graph from the beginning, excluding the part where he takes Helen's place in order to confirm his hypothesis, the inference results were the same as when he used the 90\% knowledge graph.

In other words, the reasoning for motive and opportunity was based on the description of the events leading up to the day of the crime. In contract, the reasoning for means was based on the description of Roylott's room, which Holmes first described when he revealed the seed.

\section{Future work and Summary}
In this paper we describe the application of one logical approach to crime scene investigation.
The subject of the application is ``The Adventure of the Speckled Band'' from the Sherlock Holmes short stories.
The applied data is the knowledge graph created
for the Knowledge Graph Reasoning Challenge.
We tried to find the murderer by inferring the person who has the motive, opportunity and method.
We created an ontology of motives and methods of murder from dictionaries and dictionaries, added it to the knowledge graph of ``The Adventure of the Speckled Band'', and applied scripts to determine motives, opportunities, and methods.
From these scripts, we obtained characters with motives, characters with opportunities to kill, and characters who could execute the methods of killing deduced from the situation at the crime scene. We made the final overall judgement manually based on the results of the judgments of motive, opportunity, and method.

We believe that the method used to find the murderer is applicable to murder cases in general.
However, the ontology does not describe all the motives and methods of murder.
The ontology, however, does not describe all the motives and methods of murder, so it is necessary to improve it in the future.
In this study, the overall judgment was made by hand, but it is also a future task to make it automatically.
In addition, we consider the following two points to be essential for the inference part.

\begin{itemize}
\item {\bf Distinction between ``statement'', ``thought'' and ``fact''}: 
This study, we did not analyse the suspect's words, actions, knowledge, intentions, or judgments of lies, but treated all statements and thoughts as facts.
A statement is not a fact if it is a lie.
What you think can be different from what is true.
\item {\bf Handling of time} :
To deduce who could have committed the actual crime, it is necessary to specify the crime's time, place and circumstances, and estimate who could have been there at that time (or could have caused it to happen).
First, the estimated time of the crime is described as a period.
Then, it is necessary to describe where the suspect was before and after that time.
Furthermore, the relationship between the crime scene and the places where the suspect was before and after the crime is serious, and the question is whether the suspect can physically travel to the crime scene in time.
In the present study, it was necessary to describe the location of rooms and corridors, the location of doors, windows, holes, etc., and their status (open or closed, obstruction, etc.).
\end{itemize}
The ontology, inference rules, and inference scripts for this paper are available on github\footnote{\url{https://github.com/KGChallenge/Challenge}}.



\bibliographystyle{ACM-Reference-Format}
\bibliography{lod}


\begin{thebibliography}{9}


\ifx \showCODEN    \undefined \def \showCODEN     #1{\unskip}     \fi
\ifx \showDOI      \undefined \def \showDOI       #1{#1}\fi
\ifx \showISBNx    \undefined \def \showISBNx     #1{\unskip}     \fi
\ifx \showISBNxiii \undefined \def \showISBNxiii  #1{\unskip}     \fi
\ifx \showISSN     \undefined \def \showISSN      #1{\unskip}     \fi
\ifx \showLCCN     \undefined \def \showLCCN      #1{\unskip}     \fi
\ifx \shownote     \undefined \def \shownote      #1{#1}          \fi
\ifx \showarticletitle \undefined \def \showarticletitle #1{#1}   \fi
\ifx \showURL      \undefined \def \showURL       {\relax}        \fi
\providecommand\bibfield[2]{#2}
\providecommand\bibinfo[2]{#2}
\providecommand\natexlab[1]{#1}
\providecommand\showeprint[2][]{arXiv:#2}

\bibitem[Chabot et~al\mbox{.}(2014)]%
        {CHABOT2014S95}
\bibfield{author}{\bibinfo{person}{Yoan Chabot}, \bibinfo{person}{Aur^^c3^^a9lie Bertaux}, \bibinfo{person}{Christophe Nicolle}, {and} \bibinfo{person}{M-Tahar Kechadi}.} \bibinfo{year}{2014}\natexlab{}.
\newblock \showarticletitle{A complete formalized knowledge representation model for advanced digital forensics timeline analysis}.
\newblock \bibinfo{journal}{\emph{Digital Investigation}}  \bibinfo{volume}{11} (\bibinfo{year}{2014}), \bibinfo{pages}{S95--S105}.
\newblock
\showISSN{1742-2876}
\urldef\tempurl%
\url{https://doi.org/10.1016/j.diin.2014.05.009}
\showDOI{\tempurl}
\newblock
\shownote{Fourteenth Annual DFRWS Conference}.


\bibitem[Doyle and Macaluso(2016)]%
        {doyle2016adventure}
\bibfield{author}{\bibinfo{person}{A.C. Doyle} {and} \bibinfo{person}{P.J. Macaluso}.} \bibinfo{year}{2016}\natexlab{}.
\newblock \bibinfo{booktitle}{\emph{The Adventure of the Speckled Band}}.
\newblock \bibinfo{publisher}{MX Publishing, Limited}.
\newblock
\showISBNx{9781780928838}
\urldef\tempurl%
\url{https://books.google.co.jp/books?id=qC2hCwAAQBAJ}
\showURL{%
\tempurl}


\bibitem[Elezaj et~al\mbox{.}(2019)]%
        {Elezaj2019}
\bibfield{author}{\bibinfo{person}{Ogerta Elezaj}, \bibinfo{person}{Sule Yildirim~Yayilgan}, \bibinfo{person}{Edlira Vakaj}, \bibinfo{person}{Linda Wendelberg}, \bibinfo{person}{Mohamed Abomhara}, {and} \bibinfo{person}{Javed Ahmed}.} \bibinfo{year}{2019}\natexlab{}.
\newblock \showarticletitle{Towards Designing a Knowledge Graph-Based Framework for Investigating and Preventing Crime on Online Social Networks}.
\newblock \bibinfo{journal}{\emph{Lecture Notes in Computer Science}} (\bibinfo{date}{12} \bibinfo{year}{2019}).
\newblock


\bibitem[Elsayed(2018)]%
        {Elsayed2018}
\bibfield{author}{\bibinfo{person}{Eman Elsayed}.} \bibinfo{year}{2018}\natexlab{}.
\newblock \showarticletitle{Semantic Crime Investigation System}.
\newblock \bibinfo{journal}{\emph{Journal of Computers}} (\bibinfo{date}{01} \bibinfo{year}{2018}), \bibinfo{pages}{1216--1226}.
\newblock
\urldef\tempurl%
\url{https://doi.org/10.17706/jcp.13.10.1216-1226}
\showDOI{\tempurl}


\bibitem[Jalil et~al\mbox{.}(2017)]%
        {Jalil2017}
\bibfield{author}{\bibinfo{person}{Masita Jalil}, \bibinfo{person}{Chia Ling}, \bibinfo{person}{Noor Maizura}, {and} \bibinfo{person}{Fatihah Mohd}.} \bibinfo{year}{2017}\natexlab{}.
\newblock \showarticletitle{Knowledge Representation Model for Crime Analysis}.
\newblock \bibinfo{journal}{\emph{Procedia Computer Science}}  \bibinfo{volume}{116} (\bibinfo{date}{12} \bibinfo{year}{2017}), \bibinfo{pages}{484--491}.
\newblock
\urldef\tempurl%
\url{https://doi.org/10.1016/j.procs.2017.10.067}
\showDOI{\tempurl}


\bibitem[Kadiyan(2021)]%
        {Sanskriti_Kadiyan_2021}
\bibfield{author}{\bibinfo{person}{Sanskriti Kadiyan}.} \bibinfo{year}{2021}\natexlab{}.
\newblock \showarticletitle{The Role of Artificial Intelligence in Criminal Justice System with Special Reference to Forensic Science Investigation Techniques}.
\newblock \bibinfo{journal}{\emph{Design Engineering}} (\bibinfo{date}{Oct.} \bibinfo{year}{2021}), \bibinfo{pages}{5354--5362}.
\newblock
\urldef\tempurl%
\url{http://www.thedesignengineering.com/index.php/DE/article/view/5487}
\showURL{%
\tempurl}


\bibitem[Kawamura et~al\mbox{.}(2020)]%
        {10.1007/978-3-030-41407-8_2}
\bibfield{author}{\bibinfo{person}{Takahiro Kawamura}, \bibinfo{person}{Shusaku Egami}, \bibinfo{person}{Koutarou Tamura}, \bibinfo{person}{Yasunori Hokazono}, \bibinfo{person}{Takanori Ugai}, \bibinfo{person}{Yusuke Koyanagi}, \bibinfo{person}{Fumihito Nishino}, \bibinfo{person}{Seiji Okajima}, \bibinfo{person}{Katsuhiko Murakami}, \bibinfo{person}{Kunihiko Takamatsu}, \bibinfo{person}{Aoi Sugiura}, \bibinfo{person}{Shun Shiramatsu}, \bibinfo{person}{Xiangyu Zhang}, {and} \bibinfo{person}{Kouji Kozaki}.} \bibinfo{year}{2020}\natexlab{}.
\newblock \showarticletitle{Report on the First Knowledge Graph Reasoning Challenge 2018}. In \bibinfo{booktitle}{\emph{Semantic Technology}}, \bibfield{editor}{\bibinfo{person}{Xin Wang}, \bibinfo{person}{Francesca~Alessandra Lisi}, \bibinfo{person}{Guohui Xiao}, {and} \bibinfo{person}{Elena Botoeva}} (Eds.). \bibinfo{publisher}{Springer International Publishing}, \bibinfo{address}{Cham}, \bibinfo{pages}{18--34}.
\newblock
\showISBNx{978-3-030-41407-8}


\bibitem[Srimukh and Shridevi(2020)]%
        {Srimukh2020}
\bibfield{author}{\bibinfo{person}{P~V Srimukh} {and} \bibinfo{person}{S Shridevi}.} \bibinfo{year}{2020}\natexlab{}.
\newblock \showarticletitle{Development of Ontology on Crime Investigation process}.
\newblock \bibinfo{journal}{\emph{Journal of Physics: Conference Series}}  \bibinfo{volume}{1716} (\bibinfo{date}{12} \bibinfo{year}{2020}), \bibinfo{pages}{012053}.
\newblock
\urldef\tempurl%
\url{https://doi.org/10.1088/1742-6596/1716/1/012053}
\showDOI{\tempurl}


\bibitem[Vakaj et~al\mbox{.}(2017)]%
        {Vakaj2017}
\bibfield{author}{\bibinfo{person}{Edlira Vakaj}, \bibinfo{person}{Sule Yildirim~Yayilgan}, \bibinfo{person}{Elton Domnori}, {and} \bibinfo{person}{Ogerta Elezaj}.} \bibinfo{year}{2017}\natexlab{}.
\newblock \showarticletitle{SMONT: an ontology for crime solving through social media}.
\newblock \bibinfo{journal}{\emph{International Journal of Metadata Semantics and Ontologies}}  \bibinfo{volume}{12} (\bibinfo{date}{01} \bibinfo{year}{2017}).
\newblock
\urldef\tempurl%
\url{https://doi.org/10.1504/IJMSO.2017.10011827}
\showDOI{\tempurl}


\end{thebibliography}

\end{document}